\documentclass[
]{ceurart}

\usepackage{listings}
\usepackage{comment}
\usepackage{subcaption}
\begin{document}

\copyrightyear{2026}
\copyrightclause{Copyright for this paper by its authors.
  Use permitted under Creative Commons License Attribution 4.0
  International (CC BY 4.0).}

\conference{9th International Workshop on Computational Models of Narrative (CMN'26),
  June 08--10, 2026, Universidad Complutense de Madrid, Spain}

\title{Do BERT Embeddings Encode Narrative Dimensions? A Token-Level Probing Analysis of Time, Space, Causality, and Character in Fiction}

\author{Beicheng Bei}[%
orcid=0009-0000-3869-8211,
email=bbei@ur.rochester.edu,
]

\address{Department of Linguistics, 
University of Rochester, 503 Lattimore Hall, Rochester, NY 14627}
  
\author{Hannah Hyesun Chun}[%
orcid=0009-0000-6696-5495,
email=hchun7@ur.rochester.edu,
]

\author{Chen Guo}[%
orcid=0009-0006-9111-7280,
email=cguo19@ur.rochester.edu,
]

\author{Arwa Saghiri}[%
orcid=0009-0003-8381-7357,
email=asaghiri@u.rochester.edu,
]

\begin{abstract}
Narrative understanding requires multidimensional semantic structures. This study investigates whether BERT embeddings encode dimensions of fictional narrative semantics\textemdash time, space, causality, and character. Using an LLM to accelerate annotation, we construct a token-level dataset labeled with these four narrative categories plus ``others.'' A linear probe on BERT embeddings (94\% accuracy) significantly outperforms a control probe on variance-matched random embeddings (47\%), confirming that BERT encodes meaningful narrative information. With balanced class weighting, the probe achieves a macro-average recall of 0.83, with moderate success on rare categories such as causality (recall = $0.75$) and space (recall = $0.66$). However, confusion matrix analysis reveals ``Boundary Leakage,'' where rare dimensions are systematically misclassified as ``others.'' Clustering analysis shows that unsupervised clustering aligns near-randomly with predefined categories (ARI = $0.081$), suggesting that narrative dimensions are encoded but not as discretely separable clusters. Future work includes a POS-only baseline to disentangle syntactic patterns from narrative encoding, expanded datasets, and layer-wise probing.
\end{abstract}


\maketitle

\section{Introduction}
Narrative comprehension is a fundamental aspect of human intelligence. When reading a story, we effortlessly track characters across scenes, understand temporal sequences, recognize spatial relationships, and infer causal connections between events. These narrative dimensions\textemdash character, time, space, and causality\textemdash form the conceptual scaffolding that allows us to construct coherent mental models of fictional worlds. As LLMs demonstrate increasingly sophisticated performance on natural language tasks, a critical question emerges: do these models develop internal representations that capture the narrative structure humans use to understand stories?

Recent work in model interpretability has employed probing classifiers to investigate what linguistic knowledge is encoded in pretrained representations \cite{belinkov2022probing}. Studies have examined syntactic structures, semantic roles, and various linguistic phenomena, revealing that models like BERT (Bidirectional Encoder Representations from
Transformers) \cite{devlin2019bert} encode rich syntactic information but show more limited semantic understanding \cite{tenney2019what}. However, this line of research has focused primarily on sentence-level linguistic properties rather than discourse-level narrative structure. Narrative dimensions operate at a higher level of abstraction than syntax or even sentence semantics\textemdash they organize information across multiple sentences and require tracking relationships that span entire texts.

\section{Related Work}
\subsection{Computational Narrative Analysis}
Our selection of the four narrative dimensions \textemdash character, causality, time, and space\textemdash is based on the components of narratives presented in \cite{piper2021narrative}. Although there is no unified agreement on what is classified as a narrative, Piper notes that across the various theories of narratives, all share a focus on story-level phenomena, in other words, the structure of events that unfold in a narrative. Genette\textquotesingle s narrative triangle, one of the classical models in narratology, introduces three aspects that together make up narrative: story, which refers to the events themselves; discourse, the way those events are selected and ordered; and narrating, the narrator’s role in presenting that information to weave a story \cite{genette1983narrative}. Piper presents core narrative components onto those three aspects. Agents, events, temporality, and setting correspond to the story and discourse levels, while perspective aligns with narrating.

Agent plays an important role in narrative because a narrative usually presupposes the presence of some entity that acts or perceives. Events capture the flow of actions performed by agents, showing what happens in the narrative. The sequence of events plays a critical role in identifying both temporal and logical relations between events. Logical relations serve as the foundation for causality and an improved modeling of causal relations in stories, which contributes to understanding how narratives reflect human reasoning. Temporal relations are essential to structuring events to establish a coherent plot that forms the basis of temporality, which describes when an action takes place. Setting defines where the plot unfolds and can be grounded in either the real world, the fictional world, or both. Just as agents are understood through their actions, setting influences the context in which events and agents operate. Perspective refers to the relationship between the narrator and the story, whether the narrator exists within the story world or outside of it. It shapes how events, agents, settings, and temporality are filtered and revealed in the narrative.

Our research deals with the structural aspects of narrative, defining our narrative dimensions in terms of core components drawn from story and discourse. Specifically, we adopt four dimensions: character (agents), causality (events and their relations), time (temporality), and space (setting) to annotate text data. Because our annotations are performed at the token level, we focus on aspects of narrative that are observable within a sentence or across a small number of sentences. Narrating-level components like perspective are excluded, as they often remain stable across larger segments like paragraphs and chapters. We conclude that identifying such shifts in perspective requires discourse-level analysis beyond the scope of token-level annotation.

While narrative dimensions can span entire texts in their fullest theoretical sense, our token-level annotations capture local narrative signals\textemdash the linguistic markers that serve as building blocks for larger narrative structures. For instance, we annotate explicit temporal markers (``yesterday,'' ``now'') rather than attempting to reconstruct complete timelines, and we label character mentions (``Mr. Bennet,'' ``he'') rather than tracking character development arcs. This approach allows us to investigate whether BERT's embeddings encode the fundamental categorical distinctions between narrative dimensions, even if our annotations cannot capture their full discourse-level complexity.

\subsection{LLMs for Text Annotation}
LLMs have transformed text annotation practices, enabling efficient large-scale annotation at relatively low cost. However, LLM-based annotation faces challenges including bias, inconsistency, and reliability concerns \cite{tornberg2024best}. Best practices emphasize structured prompting, clear task definitions, and rigorous validation through human review. For our work, we adopt a hybrid approach using ChatGPT \cite{singh2025openai} for initial token-level annotation guided by an iteratively refined codebook, with all 5,088 annotated tokens manually verified and corrected where necessary. This human-in-the-loop methodology balances efficiency with quality assurance.

\subsection{Probing Language Models}
Probing classifiers have emerged as a popular methodology for interpreting what linguistic and semantic information is encoded in neural language model representations. The core approach involves training a simple classifier\textemdash typically a linear model\textemdash to predict linguistic properties directly from frozen model embeddings, without fine-tuning the underlying model \cite{belinkov2022probing}. This framework allows researchers to investigate whether pretrained representations capture specific types of knowledge that may not be explicitly part of the model's training objective.

Early probing work focused on contextualized representations like ELMo (Embeddings from Language Models) \cite{peters-etal-2018-deep} and BERT, examining what syntactic and semantic phenomena these models encode. \citet{tenney2019what} introduced edge probing tasks to systematically evaluate word-level contextual representations across syntactic, semantic, local, and long-range phenomena. Their findings revealed that models trained on language modeling and translation objectives produce strong representations for syntactic structures but offer only modest improvements over non-contextual baselines for semantic tasks. This asymmetry between syntactic and semantic encoding has important implications for understanding what these models learn from self-supervised pretraining.

However, the probing paradigm faces several methodological challenges that complicate the interpretation of results. A fundamental concern is selectivity: does high probe accuracy indicate that representations genuinely encode linguistic structure, or merely that the probe has memorized patterns in the training data? \citet{hewitt2019designing} addresses this through control tasks, which pair word types with random outputs to measure a probe's memorization capacity independent of representational quality. They demonstrate that popular probes on ELMo representations lack selectivity, achieving high accuracy even on meaningless control tasks. Their work establishes that a good probe should be selective, achieving high accuracy on genuine linguistic tasks while performing poorly on controls. This insight is critical for our work, as we implement control probes trained on random embeddings to validate that our results reflect meaningful encoding of narrative dimensions rather than artifacts of high-dimensional data.

\section{Data}
The data used for this experiment is the first five chapters of \emph{Pride and
	Prejudice} by Jane Austen, originally published in 1813. The Gutenberg
Project \footnote{\url{https://www.gutenberg.org/}} provides free digital access to the complete novel in HTML
format \cite{austen1813_pride}. For this study, we extracted Chapters~1 through 5, individually saved them as a text file, and removed copyright lines and illustration captions.

To annotate the chapter, we adopted LLM-assisted annotation using ChatGPT, followed by human verification. We iteratively developed and refined a codebook, which was used as the LLM prompt. The codebook specified:

\begin{itemize}\setlength{\itemsep}{5pt}\setlength{\itemsep}{0pt}\setlength{\parskip}{0pt}
	\item \textbf{Segmentation}: Text was segmented at the sentence
	      level. The final output required accurate sentence IDs to align with
	      token embeddings.
	\item \textbf{Tokenization}: The LLM was instructed not to include
	      punctuation, except in special cases such as hyphenated compounds
	      (e.g., \emph{good-humoured}) or abbreviations (e.g., \emph{Mr.}).
	      Multi-word expressions with unified semantic function (e.g., \emph{in
	      front of}) were treated as single tokens.
	\item \textbf{Token label assignment}: Each narrative dimension was
	      explicitly defined with examples. Tokens not matching any category
	      were labeled as ``others''.
	\item \textbf{Formatting}: The output was structured as a JSONL file,
	      with each entry containing a document ID, sentence ID, token, and
	      assigned label (e.g.,
	      \texttt{{\{}"doc\_id": "Ch1", "sent\_id": 1, "token": "man", "label": "character"{\}}}).
\end{itemize}

This standardized output format supported both efficient human verification and downstream analysis.

The JSONL output contained an entry for every token in each sentence of
the chapter. We manually reviewed all tokens in context, examining them
sentence by sentence to ensure accuracy and reduce inconsistencies.
Ambiguous or misclassified tokens were discussed collaboratively and
relabeled when necessary. Because the ``others'' category constitutes a
majority of the dataset, special attention was devoted to verifying
tokens wrongly assigned to this class due to their higher risk of mislabeling.

\begin{figure}[t]
    \centering
    \begin{minipage}{\columnwidth}
        \centering
        \includegraphics[width=0.70\linewidth]{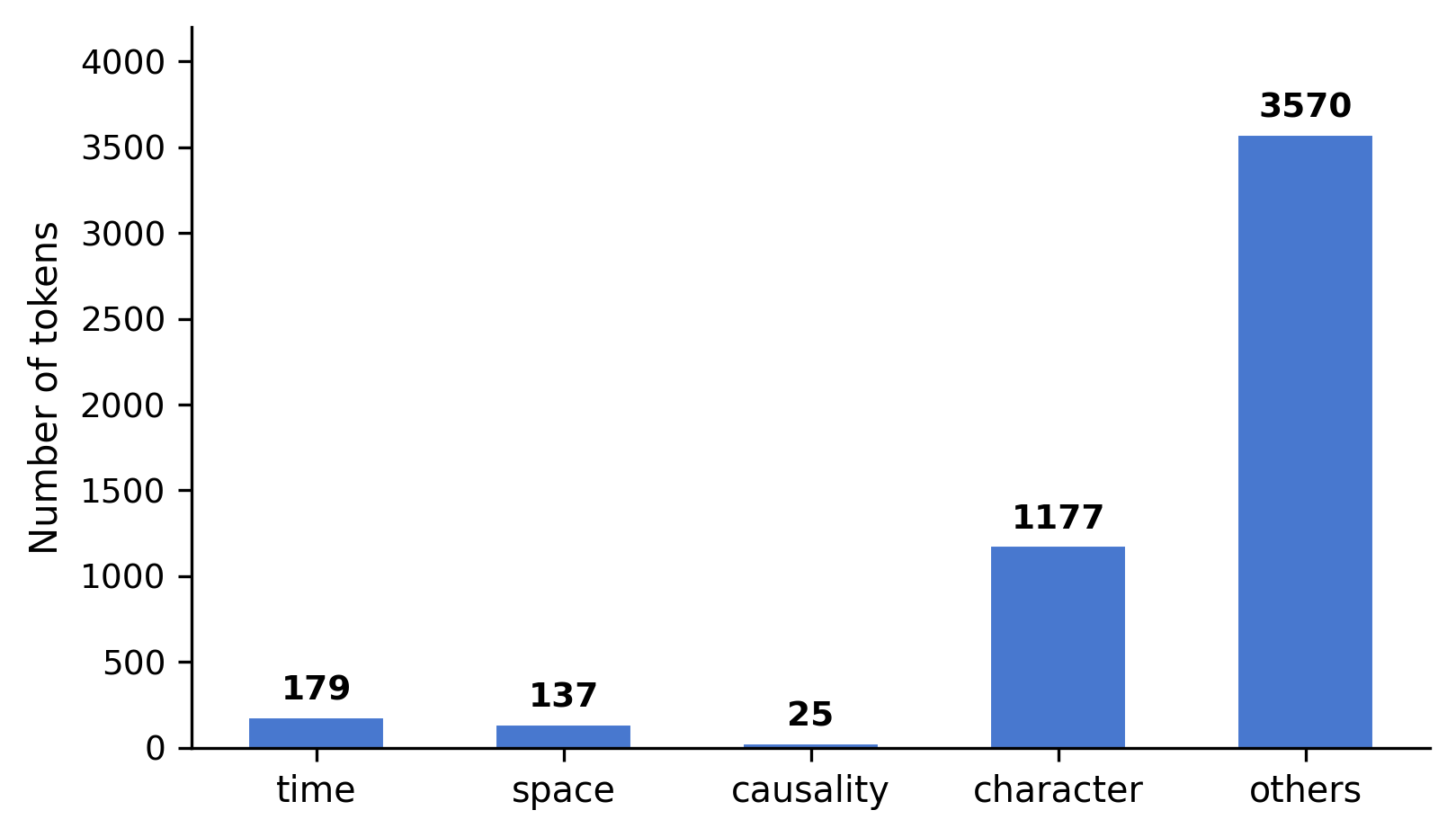}
        \caption{Label distribution in the full dataset ($n=5{,}088$). The majority class ``others'' (70.1\%) contains over 142 times more samples than the rarest class ``causality'' (0.5\%).}
        \label{fig:label-distribution}
    \end{minipage}
\end{figure}

The final dataset contains 5,088 annotated tokens for Chapters 1 through 5 of
\emph{Pride and Prejudice}. As shown in Figure~\ref{fig:label-distribution}, the dataset exhibits severe class imbalance: the ``others'' category dominates at 70.1\% of tokens, followed by ``character'' at 23.1\%. The remaining categories\textemdash ``time'' (3.5\%), ``space'' (2.7\%), and ``causality'' (0.5\%)\textemdash appear much less frequently.

\section{Methodology}

\subsection{BERT-based Embedding Extraction}
We used the original text of \textit{Pride and Prejudice} (Chapters 1-5) as input and extracted embeddings for each token using BERT \cite{devlin2019bert}. BERT is an encoder-only transformer model pretrained in a self-supervised manner on a large-scale English corpus, including English Wikipedia articles and 11,038 unpublished English books from the BookCorpus dataset \cite{zhu2015aligning}. BERT was selected for its ability to produce contextualized embeddings, which is crucial for our task since identical tokens may correspond to different narrative dimensions depending on the context. We used the \textit{bert-base-uncased}\footnote{https://huggingface.co/google-bert/bert-base-uncased} variant pre-trained on English data which consists of 12 layers each with 768-dimensional hidden states and approximately 110 million parameters.

To extract embeddings, we tokenized the text using BERT’s WordPiece tokenizer and split the tokenized sequence into segments that comply with the model’s maximum input length. Each segment was then passed through the pre-trained model in inference mode. The final-layer hidden states were extracted while discarding special tokens. This resulted in a matrix of contextualized token embeddings of size N x 768 for each chapter, where N corresponds to the number of BERT subword tokens in that chapter.

\subsection{Token-Embedding Alignment}

To obtain a single embedding for each annotated token, we aligned annotations with BERT subword embeddings. Although annotations were provided at the token level, a single annotation token could correspond to either a single word (e.g., \textit{Netherfield}) or multiple words (e.g., \textit {in the house}). To address this issue, we aligned annotated tokens to BERT subwords by tokenizing each annotated token with the same WordPiece tokenizer used for embedding extraction and matching the resulting subword sequence with the tokenized original text. Then, for each annotated token, the embeddings of the matched BERT subwords were averaged to produce a single 768-dimensional vector. This procedure was applied to Chapters 1-5, yielding an M x 768 embedding matrix for each chapter, where M is the number of annotated tokens. These aligned embeddings were used as input to the linear probe model. 

\subsection{Linear Probing}

To investigate whether BERT's internal representations encode narrative information, we employ the linear probing methodology \cite{hewitt2019designing}.
A linear probe is a simple linear classifier trained to predict task-specific labels directly from frozen model embeddings, without fine-tuning the underlying language model. This approach allows us to assess what linguistic or semantic information is already encoded in the pre-trained representations.

We use scikit-learn's \cite{pedregosa2011scikit} \texttt{LogisticRegression}  as our linear probe, which implements a multinomial logistic regression classifier.
The probe takes as input BERT's 768-dimensional contextual embeddings and outputs a probability distribution over our five narrative categories: time, space, causality, character, and others.
Formally, the probe learns a weight matrix $W \in \mathbb{R}^{768 \times 5}$ and bias vector $b \in\mathbb{R}^5$ such that:
$$P(y = c \mid x) = \frac{\exp(W_c^T x + b_c)}{\sum_{c'=1}^{5} \exp(W_{c'}^T x + b_{c'})}$$
where $x \in \mathbb{R}^{768}$ is the BERT embedding for a token, and $c$ represents one of the five narrative categories.

We split our 5,088 aligned token embedding pairs into training (70\%, $n = 3,561$) and test (30\%, $n = 1,527$) sets using stratified sampling. Stratification is critical in our experimental design because our dataset exhibits severe class imbalance (e.g., others: 3,570 samples vs.\ causality: 25 samples).
Without stratification, random splitting could result in some rare categories being entirely absent from either the training or test set, which would make evaluation meaningless. By using \texttt{stratify = y} in scikit-learn's \texttt{train\_test\_split}, we ensure that both splits preserve the original label distribution proportionally across all five categories.

We train the probe using the Limited-memory BFGS (L-BFGS) optimization algorithm \cite{liu1989lbfgs} with a maximum of 500 iterations. We set
\texttt{random\_state = 42} for reproducibility across all experiments.

\subsection{Addressing Class Imbalance}

As shown in Figure~\ref{fig:label-distribution}, our dataset exhibits severe class imbalance. This imbalance poses a significant challenge for standard machine learning algorithms, which tend to optimize for overall accuracy by focusing on majority classes while ignoring minority classes.

In standard logistic regression, the loss function is computed as the average cross-entropy loss over all training examples:
\[
	\mathcal{L} = -\frac{1}{N} \sum_{i=1}^{N} \log P(y_i \mid x_i)
\]
This formulation implicitly treats all classes equally. However, when class frequencies are highly skewed, the model can achieve low training loss by simply predicting the majority class (in our case, ``others'') for most tokens. Rare classes such as ``causality'' contribute only minimally to the overall loss, leading to category collapse.

To counteract this bias, we employ a class weighting scheme that assigns higher importance to rare classes during training. Specifically, we use
scikit-learn's \texttt{class\_weight = `balanced'} parameter, which automatically computes weights inversely proportional to class frequencies:
\[
	w_c = \frac{N}{K \cdot N_c}
\]
where $w_c$ is the weight for class $c$, $N$ is the total number of training samples, $K$ is the number of classes (5 in our case), and $N_c$ is the number of training samples in class $c$.

With these weights, the modified loss function becomes:
\[
	\mathcal{L}_{\text{weighted}} = -\frac{1}{N}
	\sum_{i=1}^{N} w_{y_i} \log P(y_i \mid x_i)
\]
This reweighting scheme ensures that misclassifying a rare class token (e.g., ``causality'') incurs a much larger penalty than misclassifying a common class token (e.g., ``others''), thereby incentivizing the model to pay attention to all categories.

In our experiments, applying
\texttt{class\_weight = `balanced'} yields the following weights for each category:
\begin{itemize}\setlength{\itemsep}{0pt}\setlength{\itemsep}{0pt}\setlength{\parskip}{0pt}
	\item Time (class 0): 5.70
	\item Space (class 1): 7.42
	\item Causality (class 2): 41.89
	\item Character (class 3): 0.86
	\item Others (class 4): 0.28
\end{itemize}

Note that ``causality,'' the rarest class with only 17 training samples, receives a weight approximately 150 times larger than ``others'' (41.89 vs.\ 0.28).
This dramatic reweighting forces the optimization algorithm to treat rare class predictions as critical.

Class weighting addresses the optimization bias in imbalanced learning, but cannot overcome fundamental data scarcity. While we expect improved recall and F1 scores for underrepresented categories like time and space (which have 125 and 96 training samples respectively), we anticipate that causality---with only 17 training examples---may still prove difficult for the probe to learn. This hypothesis will be evaluated in our results section.

\subsection{Control Probe}

A critical question arises when interpreting probing results: how do we know that the probe's performance reflects meaningful information encoded in BERT's representations, rather than artifacts of the data or spurious patterns that any high-dimensional vector space might exhibit?
To address this concern, we implement a control probe experiment following practices established in the probing literature \cite{hewitt2019designing}. 

The control probe methodology tests the null hypothesis that a linear classifier could achieve similar performance on random embeddings as on real BERT embeddings. If this null hypothesis holds, it would suggest that the probe's success is not due to meaningful linguistic structure in BERT's representations, but rather to superficial patterns or the probe's ability to memorize the training data.

To implement this control, we generate random embeddings that match the dimensionality and distributional characteristics of real BERT embeddings but contain no linguistic information. Specifically:

\begin{itemize}
	\item We generate random embeddings with identical dimensions to our
	      real data:
	      \[
		      X_{\text{train}}^{\text{random}} \in \mathbb{R}^{3561 \times 768},
		      \qquad
		      X_{\text{test}}^{\text{random}} \in \mathbb{R}^{1527 \times 768}.
	      \]

	\item We sample from a normal distribution
	      $\mathcal{N}(0, \sigma_{\text{BERT}}^2)$, where
	      $\sigma_{\text{BERT}} = 0.5355$ is the standard deviation of the real
	      BERT embeddings. This ensures the random baseline matches the
	      variance of real embeddings, so that any performance difference
	      reflects information content rather than distributional mismatch.

	\item We train the control probe using exactly the same
	      hyperparameters as the real probe:
	      \begin{itemize}
		      \item Model: \texttt{LogisticRegression} with L-BFGS solver
		      \item Maximum iterations: 500
		      \item Class weights: \texttt{`balanced'}
		      \item Random seed: 42 (reproducibility)
	      \end{itemize}
\end{itemize}

Crucially, we use the same $y_{\text{train}}$ and $y_{\text{test}}$ labels that were used with real embeddings. This ensures that any performance difference can be attributed to the information content of the embeddings, rather than differences in the supervision signal.

\begin{figure*}[t]
    \centering
    \begin{subfigure}[t]{0.48\textwidth}
        \centering
        \includegraphics[width=\linewidth]{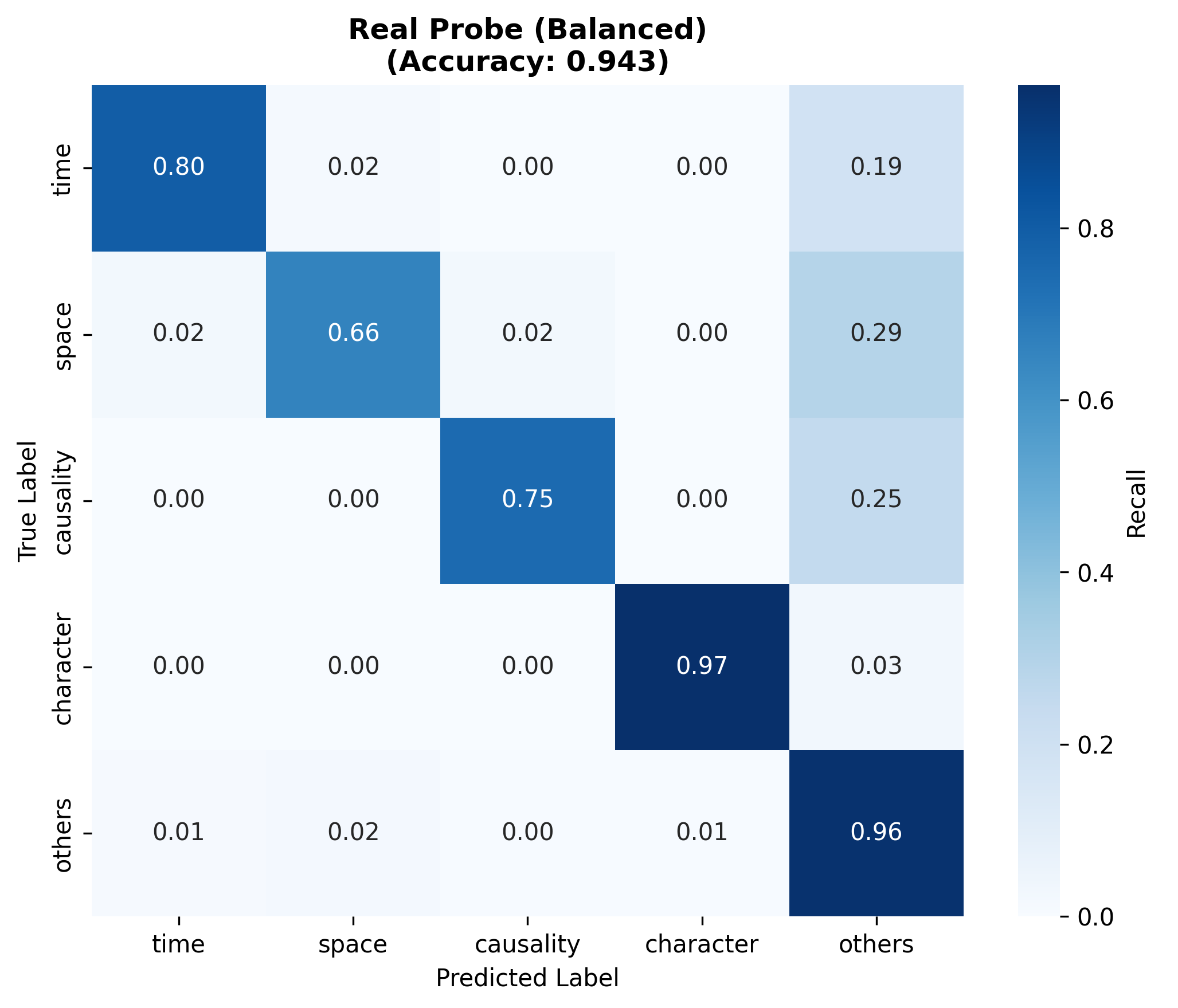}
        \caption{Real Probe (Accuracy: 94\%)}
        \label{fig:probe-real}
    \end{subfigure}
    \hfill
    \begin{subfigure}[t]{0.48\textwidth}
        \centering
        \includegraphics[width=\linewidth]{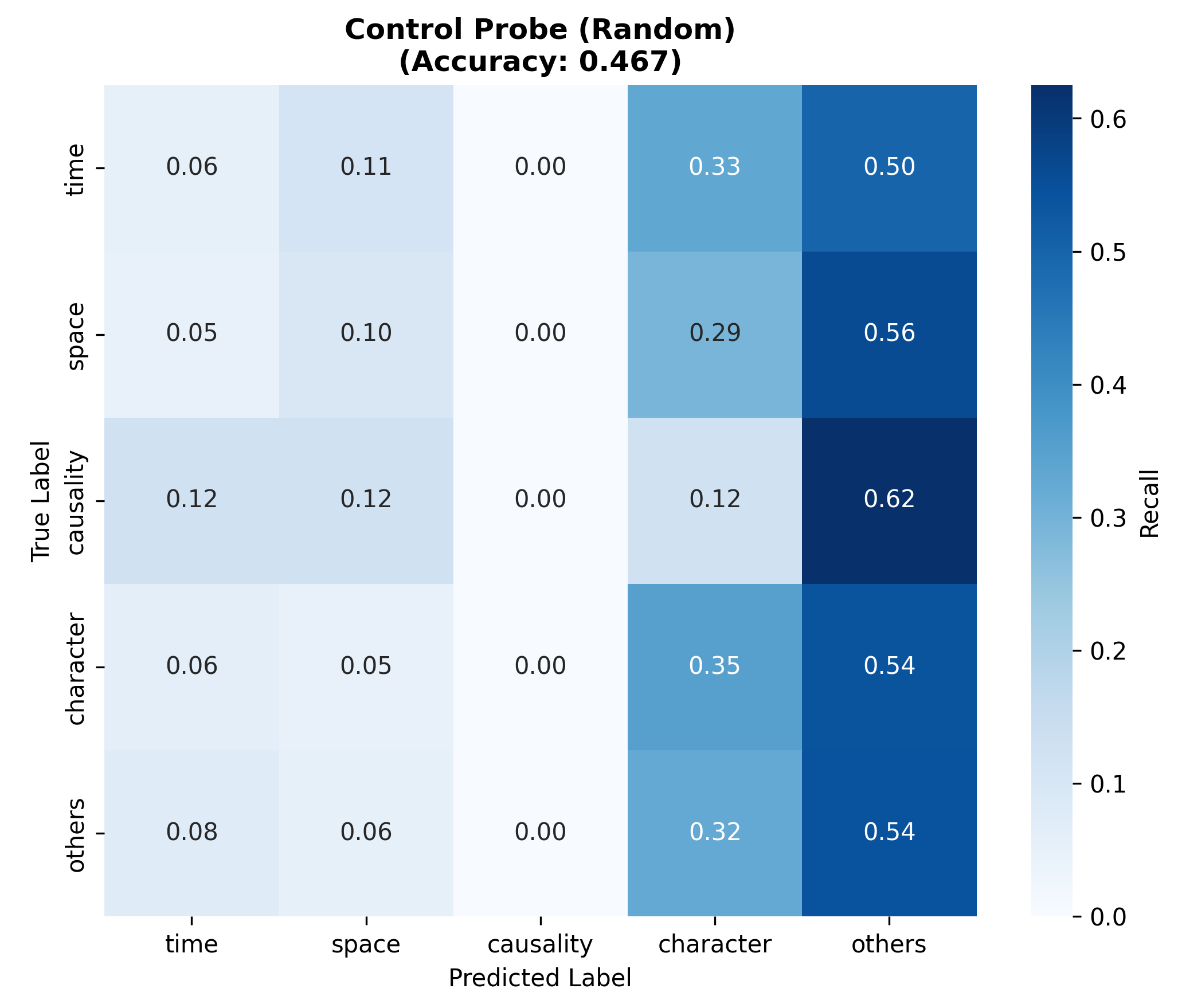}
        \caption{Control Probe (Accuracy: 47\%)}
        \label{fig:probe-control}
    \end{subfigure}
    \caption{Confusion matrices for the real probe (a) and control probe (b), both trained with balanced class weights.}
    \label{fig:probe}
\end{figure*}

\begin{table}[t]
    \centering
    \begin{minipage}{0.8\columnwidth}
        \centering
        \small
        \begin{tabular}{lcccc}
        \hline
        \textbf{Class} & \textbf{Precision} & \textbf{Recall} & \textbf{F1} & \textbf{Support} \\
        \hline
        time       & 0.75 & 0.80 & 0.77 & 54 \\
        space      & 0.53 & 0.66 & 0.59 & 41 \\
        causality  & 0.50 & 0.75 & 0.60 & 8 \\
        character  & 0.98 & 0.97 & 0.97 & 353 \\
        others     & 0.97 & 0.96 & 0.96 & 1,071 \\
        \hline
        macro avg  & 0.75 & 0.83 & 0.78 & 1,527 \\
        weighted avg & 0.95 & 0.94 & 0.94 & 1,527 \\
        \hline
        \end{tabular}
        \caption{Classification report for the real probe with balanced class weights on the Ch1--5 test set ($n = 1{,}527$).}
        \label{tab:metric}
    \end{minipage}
\end{table}

\section{Results}
The real probe, trained with balanced class weights on BERT embeddings, achieved an overall accuracy of 94\%, outperforming the control probe at 47\% (Figure~\ref{fig:probe}). 

A per-class examination (Table~\ref{tab:metric}) reveals that performance varies substantially across narrative dimensions. The majority classes---``character'' (F1 = 0.97) and ``others'' (F1 = 0.96)---are classified with high reliability. The rare classes show more moderate but meaningful performance: ``time'' achieves an F1 of 0.77 (precision = 0.75, recall = 0.80), ``causality'' an F1 of 0.60 (precision = 0.50, recall = 0.75), and ``space'' an F1 of 0.59 (precision = 0.53, recall = 0.66). The macro-average recall of 0.83 indicates that, with balanced class weighting, the probe correctly identifies the majority of positive instances across all categories.

The causality class warrants closer examination given its small sample size ($n=25$, with only 17 training instances). As shown in Table~\ref{tab:causality}, the 25 causality tokens comprise only 7 unique word types, dominated by causal connectives that function as subordinating conjunctions, adverbs, or prepositions signaling causal or consequential relations between narrative events. The probe's F1 of 0.60 on this class, while above the control baseline, should be interpreted with caution: the limited lexical diversity and extremely small sample size mean that the probe may be learning token-specific patterns rather than a general representation of narrative causality. We note that causal connectives like \emph{for} are highly ambiguous---the same word is labeled as causality in some contexts and others elsewhere---which makes the probe's ability to distinguish these uses noteworthy, though not conclusive given the sample size.

\begin{table}[t]
    \centering
    \begin{minipage}{0.85\columnwidth}
        \centering
        \small
        \begin{tabular}{l c p{5cm}}
        \hline
        \textbf{Token} & \textbf{Count} & \textbf{Example context} \\
        \hline
        for & 11 & ``\ldots\ impossible \textbf{for} her to introduce him, for she will not know him'' \\
        therefore & 6 & ``\ldots\ and \textbf{therefore} you must visit him as soon as he comes'' \\
        because & 3 & ``His pride does not offend me \ldots\ \textbf{because} there is an excuse for it'' \\
        since & 2 & ``It will be no use \ldots\ \textbf{since} you will not visit them'' \\
        consequently & 1 & ``\ldots\ and \textbf{consequently} unable to accept the honour'' \\
        thereby & 1 & ``\ldots\ His character is \textbf{thereby} complete'' \\
        so & 1 & ``\ldots\ \textbf{So} he inquired who she was'' \\
        \hline
        \textbf{Total} & \textbf{25} & \\
        \hline
        \end{tabular}
        \caption{All causality tokens in the dataset (Ch.\ 1--5 of \emph{Pride and Prejudice}), grouped by word type with frequency and example context.}
        \label{tab:causality}
    \end{minipage}
\end{table}

Analysis of the confusion matrix (Figure~\ref{fig:probe}) reveals a pattern we term ``Boundary Leakage.'' When the model misclassifies tokens from rare dimensions, it does not confuse them with each other (e.g., mistaking time for space); rather, it consistently defaults to the ``others'' class. Specifically, 19\% of time tokens, 29\% of space tokens, and 25\% of causality tokens are misclassified as ``others.'' This indicates that the ``others'' label acts as a semantic sink: when contextual cues are weak or ambiguous, the model defaults to the majority class despite balanced class weighting partially mitigating this tendency.

By contrast, the control probe achieves near-chance performance on all narrative dimensions (F1 $\leq$ 0.07 for time, space, and causality; macro-average F1 = 0.20).

\section{Analysis}
\subsection{Dimensionality Reduction and Clustering}

To visualize the distribution and relationships among the four narrative dimensions within the BERT latent space, we compared three dimensionality reduction techniques: Principal Component Analysis (PCA), Uniform Manifold Approximation and Projection (UMAP) \cite{mcinnes2018umap}, and Isometric Mapping (Isomap). Each method was used to project the high-dimensional embeddings into a two-dimensional space. The majority ``others'' class was excluded from these plots to prevent its overwhelming volume from occluding the minority dimensions.

\begin{figure*}[t]
    \centering
    \begin{subfigure}[t]{0.48\textwidth}
        \centering
        \includegraphics[width=\linewidth, height=0.7\linewidth]{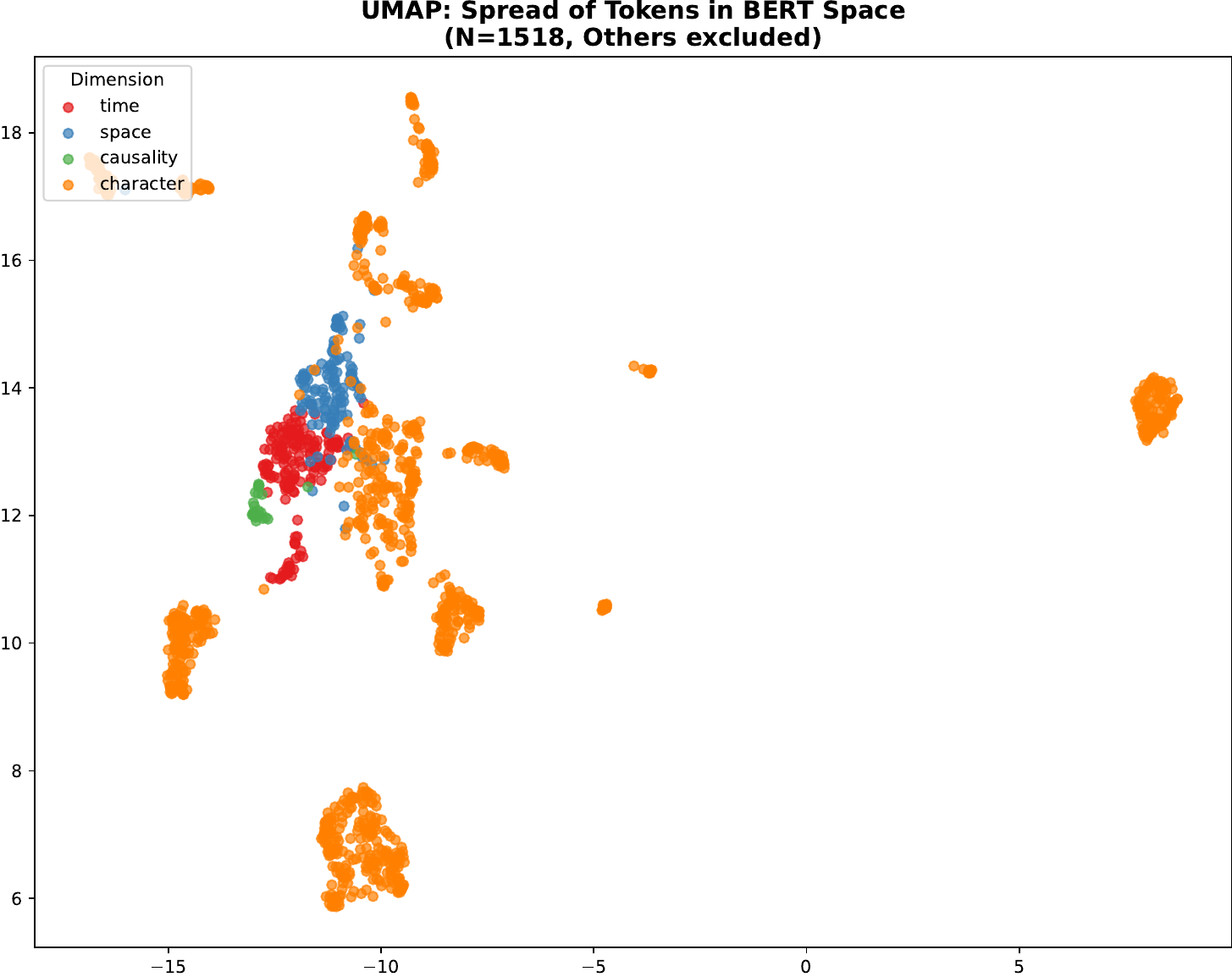}
        \caption{UMAP projection colored by narrative dimension ($N=1,518$, Others excluded).}
        \label{fig:umap_spread}
    \end{subfigure}
    \hfill
    \begin{subfigure}[t]{0.48\textwidth}
        \centering
        \includegraphics[width=\linewidth, height=0.7\linewidth]{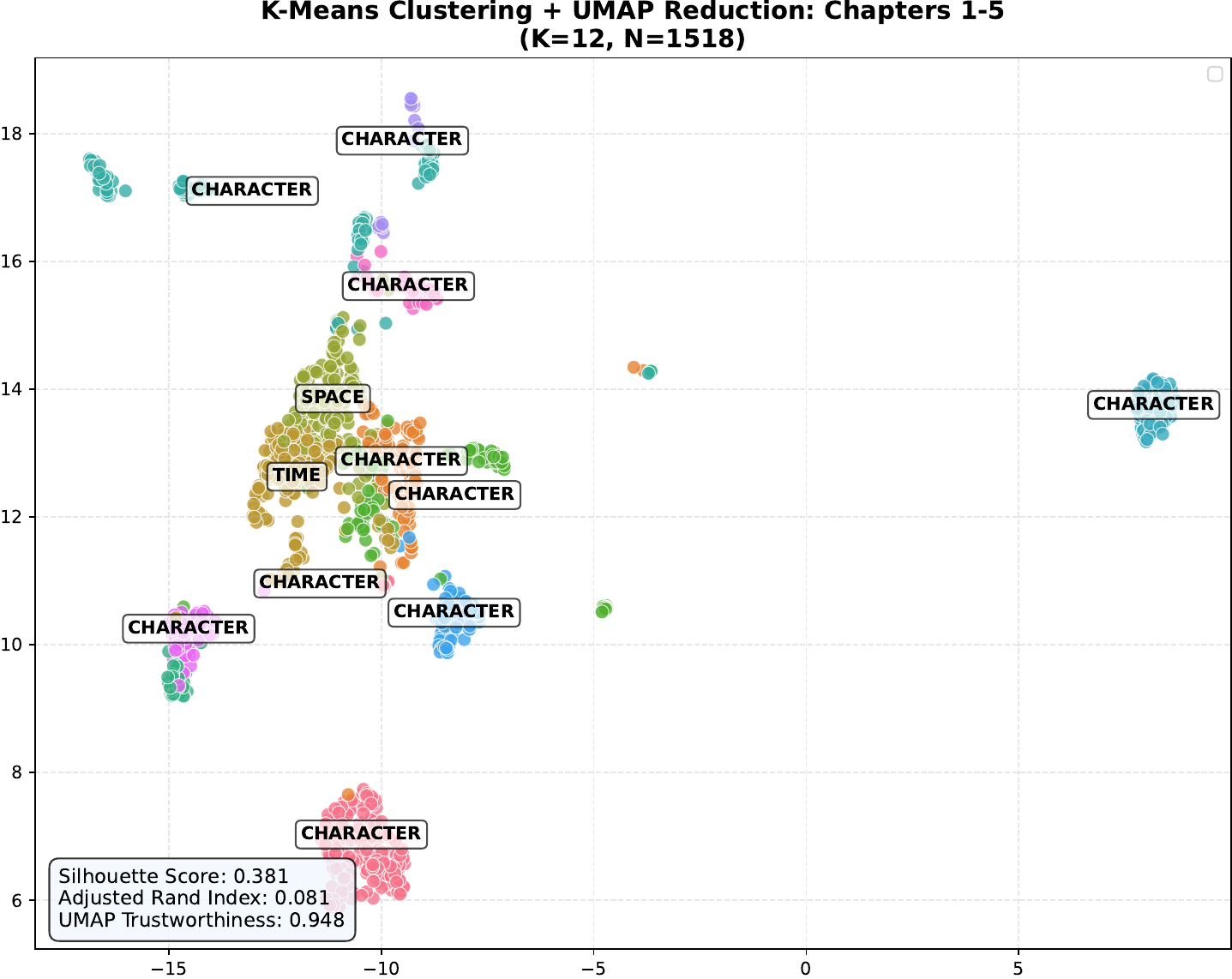}
        \caption{K-Means (768-dim, $k{=}12$) clusters visualized on UMAP projection (Silhouette=$0.381$, ARI=$0.081$).}
        \label{fig:umap_kmeans}
    \end{subfigure}
    \caption{UMAP projection and K-Means (768-dim) clustering of BERT embeddings for narrative tokens.}
\end{figure*}

The UMAP projection shows that character-related data points spread toward the periphery while other narrative dimensions gravitate toward the center (Figure~\ref{fig:umap_spread}), consistent with observations from PCA and Isomap. UMAP was selected as the primary visualization method as it better preserves the global structure of the high-dimensional embedding space.

PCA revealed that the first two principal components account for only 13.7\% of the total variance (PC1: 7.2\%, PC2: 6.5\%), indicating that a linear projection is insufficient to separate the complex narrative features encoded in BERT's 768-dimensional space. Isomap, a non-linear method that preserves geodesic distances, produced a similar spatial pattern to UMAP but with less distinct boundaries between minority categories. A UMAP Trustworthiness score of 0.948 validates that the two-dimensional projection maintains high fidelity to the original high-dimensional BERT embedding space.

We applied K-Means clustering ($k{=}12$) directly in the 768-dimensional BERT embedding space---prior to any dimensionality reduction---to examine the intrinsic structure of the embeddings. The UMAP projection serves solely as a visualization tool, with data points colored by the cluster assignments derived from the high-dimensional space (Figure~\ref{fig:umap_kmeans}). The clustering achieved a Silhouette Score of $0.381$ in the 768-dimensional space. The low Adjusted Rand Index (ARI $= 0.081$) indicates that the resulting clusters align near-randomly with the predefined narrative categories, suggesting that narrative dimensions occupy highly overlapping, non-spherical regions in BERT's embedding space rather than forming separable clusters.

The ``character'' class, which accounts for 77.54\% of tokens among the four narrative dimensions, heavily biased centroid placement, resulting in its fragmentation across multiple distinct clusters rather than forming a single semantic grouping. Notably, 11 of the 12 K-Means clusters were dominated by Character tokens, with only one cluster corresponding to the Time dimension; no dedicated Space or Causality clusters emerged. This fragmentation highlights a fundamental limitation of distance-based clustering when applied to data characterized by significant overlap and severe class imbalance. BERT embeddings are known to exhibit anisotropy and complex, overlapping semantic boundaries \cite{ethayarajh-2019-contextual}, making simple distance-based clustering insufficient for separating narrative dimensions.

\subsection{POS Distribution and Token Span}

\begin{figure}[t]
    \centering
    \begin{minipage}{0.7\linewidth}
        \centering
        \includegraphics[width=\linewidth]{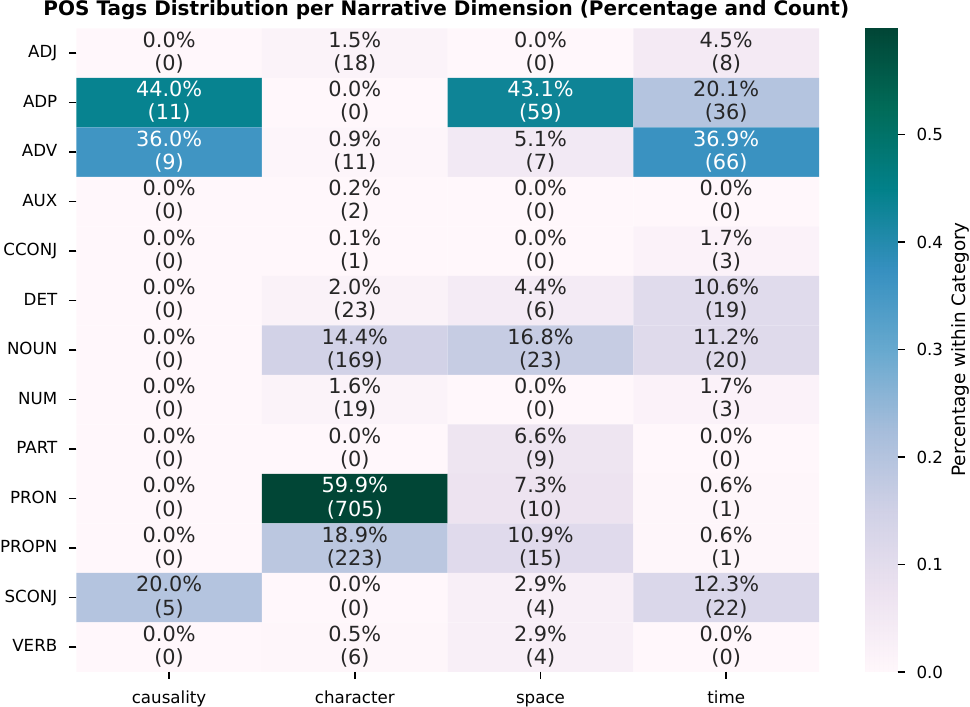}
        \caption{POS tag distribution per narrative dimension.}
        \label{fig:pos}
    \end{minipage}
\end{figure}

\begin{figure}[t]
    \centering
    \begin{minipage}{0.6\linewidth}
        \centering
        \includegraphics[width=\linewidth]{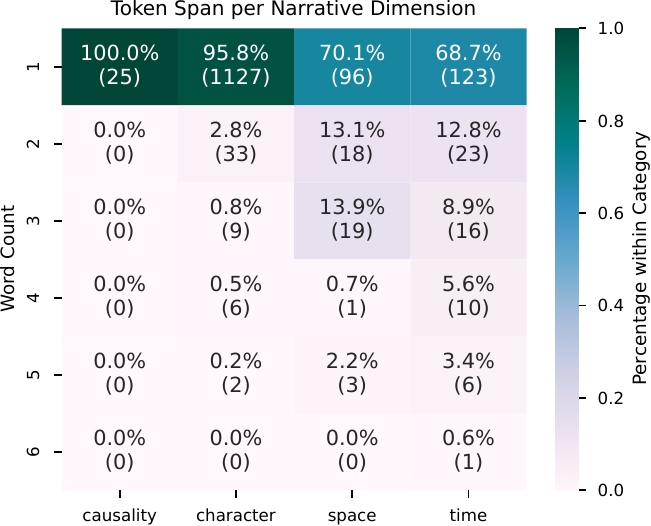}
        \caption{Token span distribution per narrative dimension.}
        \label{fig:token}
    \end{minipage}
\end{figure}

To examine the link between narrative dimensions and syntax, we analyzed the distribution of POS tags across the four narrative dimensions. Each category displays a distinct syntactic profile (Figure~\ref{fig:pos}). The \textit{Character} dimension is dominated by pronouns (PRON, 59.9\%) and proper nouns (PROPN, 18.9\%), consistent with its role in identifying narrative agents. Notably, \textit{Time} and \textit{Causality} share similar parts of speech---subordinating conjunctions (SCONJ), adverbs (ADV), and adpositions (ADP)---except that \textit{Time} is also prominently represented in determiners (DET, 10.6\%) and nouns (NOUN, 11.2\%). \textit{Space} is dominated by adpositions (ADP, 43.1\%), followed by nouns (NOUN, 16.8\%) and proper nouns (PROPN, 10.9\%).

A notable example is \textit{twenty}, which is tagged as ``character'' when it refers to a group of people (e.g., \textit{when there are twenty, I will visit them all}). Although such tokens function syntactically as determiners or adjectives, they serve a semantic role as referential nouns. This mismatch between syntactic form and narrative function highlights the challenge of performing classification purely at the semantic level.

We additionally analyzed token span lengths across narrative dimensions (Figure~\ref{fig:token}). \textit{Causality} consists entirely of single-word tokens (100\%), and \textit{Character} is predominantly single-word (95.8\%). In contrast, \textit{Space} and \textit{Time} display a wider range of token lengths: only 70.1\% of \textit{Space} tokens and 68.7\% of \textit{Time} tokens are single-word, with substantial proportions of two-word spans (\textit{Space}: 13.1\%, \textit{Time}: 12.8\%), three-word spans (\textit{Space}: 13.9\%, \textit{Time}: 8.9\%), and beyond. Because \textit{Space} and \textit{Time} involve longer token spans, they occupy a higher percentage of ADP tags with introductory words like ``for,'' ``on,'' ``in,'' and ``after.''

This multi-word nature connects directly to the Boundary Leakage pattern observed in Figure~\ref{fig:probe-real}: longer spans present greater challenges for token-level embedding alignment, potentially contributing to the higher misclassification rates of \textit{Space} (29\%) and \textit{Time} (19\%) into the ``others'' category. Furthermore, the syntactic variability of \textit{Character} tokens helps explain their broad dispersion in embedding space: because BERT builds semantic representations from low-level syntax \cite{jawahar2019what}, the diverse syntactic origins of \textit{Character} tokens lead to wider spread in all dimensionality reduction visualizations. This raises further questions about which linguistic features---beyond surface POS---BERT uses to encode narrative dimensions.

We note, however, that the strong POS--narrative correlations observed here are partly a consequence of the annotation scheme itself: by defining narrative categories in terms of linguistic function (e.g., pronouns $\rightarrow$ \textit{Character}, adpositions $\rightarrow$ \textit{Space}), the codebook introduces a systematic link between POS and label that is independent of what BERT learns. Disentangling BERT's contribution from this annotation-induced POS signal would require a controlled POS-only baseline classifier; we leave this comparison to future work.

\section{Discussion}
The primary limitation of this project is the inherent noise and label imbalance within the dataset.
These issues stem from the complexity of language itself,
as narrative semantics are not always encapsulated by single lexical units. This leads to three specific challenges as the below.
\subsection{Granularity and Tokenization}
Determining the appropriate level of granularity is non-trivial.
Narrative features often span phrases rather than individual words.
In our dataset, this leads to conflicting tokenization strategies.
For instance, the phrase ``into the neighbourhood'' is treated inconsistently: sometimes the entire phrase is labeled as \textit{Space}, while in another case, it is segmented, with ``into'' and ``neighbourhood'' labeled individually.
\subsection{Lexical Ambiguity and Inconsistency}
Polysemy (words with multiple meanings) introduces significant labeling noise. Prepositions such as ``into'' and ``by'' are particularly ambiguous. For example, ``into'' can denote physical motion or abstract investigation (e.g., ``look into'').

We observed instances where identical syntactic structures involving the word ``for'' received different dimension labels, confirming a lack of annotation consistency. For example, the phrase ``what an establishment it would be \textit{for} one of them'' is labeled as \textit{Causality}, whereas the syntactically parallel phrase ``What a fine thing \textit{for} our girls'' is labeled as \textit{Others}.
\subsection{Domain Shift}
Lastly, the text used as data presents a limitation. As the examples above illustrate, many sentences appear awkward or syntactically unconventional relative to modern usage. The English style of 1813 (\textit{Pride and Prejudice}) differs significantly from contemporary writing, and this divergence adds extra difficulty when resolving stylistic differences at the level of linguistic definition.

\section{Conclusion and Future Work}
This study investigated whether BERT embeddings encode narrative dimensions---time, space, causality, and character---using linear probing classifiers on 5,088 annotated tokens from Chapters 1--5 of \textit{Pride and Prejudice}. The real probe achieved 94\% accuracy, outperforming the control probe at 47\%, providing evidence that BERT's contextualized representations encode meaningful narrative information. With balanced class weighting, the probe achieves a macro-average recall of 0.83, demonstrating moderate success even on rare categories.

However, our analysis reveals persistent challenges. Confusion matrix analysis identifies ``Boundary Leakage,'' where rare narrative dimensions are systematically misclassified as the majority class ``others'' rather than being confused with each other. Dimensionality reduction and clustering analysis---applying K-Means in the 768-dimensional embedding space and visualizing results with UMAP---consistently shows that narrative categories are more fluid and intermixed than discrete labels suggest, with a low ARI ($0.081$) indicating that unsupervised clustering aligns near-randomly with predefined categories. POS distribution and token span analysis further reveal that multi-word narrative expressions, particularly in \textit{Space} and \textit{Time}, present challenges for token-level classification.

Based on these findings, we propose several directions for future research. First, we aim to establish more robust definitions of narrative dimensions and annotation guidelines to reduce ambiguities in semantic boundaries. Second, we plan to expand the dataset by incorporating additional literary texts, including complete short stories that contain full narrative arcs, as well as modern writing. Using complete texts rather than partial chapters of a novel may allow all narrative dimensions---particularly ``causality'', which tends to intensify in later plot development---to be more adequately represented. Third, a pivotal objective is to investigate how BERT's internal layers represent narrative features through layer-wise probing, since narrative dimensions are highly semantic and may challenge deeper layer representations more than surface-level linguistic tasks. Additionally, future work should include a POS-only baseline classifier to disentangle the contribution of surface syntactic patterns from deeper semantic encoding in BERT representations, directly addressing the confound introduced by POS-correlated annotation guidelines. Finally, we seek to apply non-linear classifiers capable of capturing the complex, anisotropic embedding structures inherent to narrative semantics, moving beyond the linear probing framework to better resolve the overlapping boundaries between narrative dimensions.

\bibliography{sample-ceur}

@String{Computer = "{IEEE} Computer" }

@inproceedings{piper2021narrative,
  title     = {Narrative Theory for Computational Narrative Understanding},
  author    = {Piper, Andrew and So, Richard Jean and Bamman, David},
  booktitle = {Proceedings of the 2021 Conference on Empirical Methods in Natural Language Processing},
  pages     = {298--311},
  year      = {2021},
  address   = {Online and Punta Cana, Dominican Republic},
  publisher = {Association for Computational Linguistics}
}

@book{genette1983narrative,
    title     = {Narrative Discourse: An Essay in Method},
    author    = {Genette, G{\'e}rard},
    year      = {1983},
    volume    = {3},
    publisher = {Cornell University Press}
}

@article{tornberg2024best,
  title   = {Best Practices for Text Annotation with Large Language Models},
  author  = {T{\"o}rnberg, Petter},
  journal = {arXiv preprint arXiv:2402.05129},
  year    = {2024},
  url     = {https://doi.org/10.48550/arXiv.2402.05129}
}

@article{singh2025openai,
  title={Openai gpt-5 system card},
  author={Singh, Aaditya and Fry, Adam and Perelman, Adam and Tart, Adam and Ganesh, Adi and El-Kishky, Ahmed and McLaughlin, Aidan and Low, Aiden and Ostrow, AJ and Ananthram, Akhila and others},
  journal={arXiv preprint arXiv:2601.03267},
  year={2025}
}

@article{belinkov2022probing,
  title   = {Probing Classifiers: Promises, Shortcomings, and Advances},
  author  = {Belinkov, Yonatan},
  journal = {Computational Linguistics},
  volume  = {48},
  number  = {1},
  pages   = {207--219},
  year    = {2022}
}

@article{tenney2019what,
  title     = {What do You Learn from Context? Probing for Sentence Structure in Contextualized Word Representations},
  author    = {Tenney, Ian and Das, Dipanjan and Pavlick, Ellie},
  journal   = {arXiv preprint arXiv:1905.06316},
  year      = {2019},
  url       = {https://arxiv.org/abs/1905.06316}
}

@inproceedings{hewitt2019designing,
  title     = {Designing and Interpreting Probes with Control Tasks},
  author    = {Hewitt, John and Liang, Percy},
  booktitle = {Proceedings of the 2019 Conference on Empirical Methods in Natural Language Processing and the 9th International Joint Conference on Natural Language Processing (EMNLP-IJCNLP)},
  pages     = {2733--2743},
  year      = {2019},
  publisher = {Association for Computational Linguistics}
}

@inproceedings{devlin2019bert,
  title={Bert: Pre-training of deep bidirectional transformers for language understanding},
  author={Devlin, Jacob and Chang, Ming-Wei and Lee, Kenton and Toutanova, Kristina},
  booktitle={Proceedings of the 2019 conference of the North American chapter of the association for computational linguistics: human language technologies, volume 1 (long and short papers)},
  pages={4171--4186},
  year={2019}
}

@inproceedings{peters-etal-2018-deep,
    title = "Deep Contextualized Word Representations",
    author = "Peters, Matthew E.  and
      Neumann, Mark  and
      Iyyer, Mohit  and
      Gardner, Matt  and
      Clark, Christopher  and
      Lee, Kenton  and
      Zettlemoyer, Luke",
    editor = "Walker, Marilyn  and
      Ji, Heng  and
      Stent, Amanda",
    booktitle = "Proceedings of the 2018 Conference of the North {A}merican Chapter of the Association for Computational Linguistics: Human Language Technologies, Volume 1 (Long Papers)",
    month = jun,
    year = "2018",
    address = "New Orleans, Louisiana",
    publisher = "Association for Computational Linguistics",
    url = "https://aclanthology.org/N18-1202/",
    doi = "10.18653/v1/N18-1202",
    pages = "2227--2237",
}

@inproceedings{ethayarajh-2019-contextual,
    title = "How Contextual are Contextualized Word Representations? Comparing the Geometry of {BERT}, {ELM}o, and {GPT}-2 Embeddings",
    author = "Ethayarajh, Kawin",
    booktitle = "Proceedings of the 2019 Conference on Empirical Methods in Natural Language Processing and the 9th International Joint Conference on Natural Language Processing (EMNLP-IJCNLP)",
    month = nov,
    year = "2019",
    address = "Hong Kong, China",
    publisher = "Association for Computational Linguistics",
    url = "https://aclanthology.org/D19-1006",
    doi = "10.18653/v1/D19-1006",
    pages = "55--65",
}

@inproceedings{jawahar2019what,
  title     = {What Does {BERT} Learn about the Structure of Language?},
  author    = {Jawahar, Ganesh and Sagot, Beno{\^\i}t and Sautier, Djam{\'e}},
  booktitle = {Proceedings of the 57th Annual Meeting of the Association for Computational Linguistics},
  pages     = {3651--3657},
  year      = {2019},
  address   = {Florence, Italy},
  publisher = {Association for Computational Linguistics}
}

@article{pedregosa2011scikit,
  title     = {Scikit-learn: Machine Learning in {P}ython},
  author    = {Pedregosa, F. and Varoquaux, G. and Gramfort, A. and Michel, V. and Thirion, B. and Grisel, O. and Blondel, M. and Prettenhofer, P. and Weiss, R. and Dubourg, V. and Vanderplas, J. and Passos, A. and Cournapeau, D. and Brucher, M. and Perrot, M. and Duchesnay, E.},
  journal   = {Journal of Machine Learning Research},
  volume    = {12},
  pages     = {2825--2830},
  year      = {2011}
}

@article{mcinnes2018umap,
  title   = {{UMAP}: Uniform Manifold Approximation and Projection for Dimension Reduction},
  author  = {McInnes, Leland and Healy, John and Melville, James},
  journal = {arXiv preprint arXiv:1802.03426},
  year    = {2018}
}

@inproceedings{zhu2015aligning,
  title     = {Aligning Books and Movies: Towards Story-Like Visual Explanations by Watching Movies and Reading Books},
  author    = {Zhu, Yukun and Kiros, Ryan and Zemel, Rich and Salakhutdinov, Ruslan and Urtasun, Raquel and Torralba, Antonio and Fidler, Sanja},
  booktitle = {Proceedings of the IEEE International Conference on Computer Vision},
  pages     = {19--27},
  year      = {2015}
}

@article{liu1989lbfgs,
  title   = {On the Limited Memory {BFGS} Method for Large Scale Optimization},
  author  = {Liu, Dong C. and Nocedal, Jorge},
  journal = {Mathematical Programming},
  volume  = {45},
  number  = {1},
  pages   = {503--528},
  year    = {1989}
}

@misc{austen1813_pride,
  author       = {Austen, Jane},
  title        = {Pride and Prejudice},
  howpublished = {Project Gutenberg EBook No. 1342, HTML edition},
  year         = {1813},
  url          = {https://www.gutenberg.org/files/1342/1342-h/1342-h.htm},
  note         = {Accessed: 10 Dec 2025}
}

\appendix

\end{document}